\documentclass[runningheads]{llncs}

\usepackage{eccv}

\usepackage{eccvabbrv}

\usepackage{graphicx}
\usepackage{booktabs}
\usepackage{multirow}
\usepackage{colortbl}
\usepackage{xspace}
\usepackage{float}
\usepackage{mathtools}
\usepackage{listings}
\usepackage[accsupp]{axessibility}  % Improves PDF readability for those with disabilities.

\usepackage{wrapfig}

\usepackage{hyperref}

\usepackage{orcidlink}

\def\ModelName{{\textcolor{orange}{Omni}\textcolor{violet}{ACT}}\normalfont}

\def\pyautogui{PyAutoGUI}

\let\svthefootnote\thefootnote
\newcommand\freefootnote[1]{%
  \let\thefootnote\relax%
  \footnotetext{#1}%
  \let\thefootnote\svthefootnote%
}

\begin{document}

% ---------------------------------------------------------------
%  REVIEW: Replace with your title
\title{\ModelName{}: A Dataset and Benchmark for Enabling Multimodal Generalist Autonomous Agents for Desktop and Web} 

% %  REVIEW: If the paper title is too long for the running head, you can set
% % an abbreviated paper title here. If not, comment out.
\titlerunning{OmniACT}

% % TODO FINAL: Replace with your author list. 
% % Include the authors' OCRID for the camera-ready version, if at all possible.
% % \author{{Raghav Kapoor\textsuperscript{$\dagger$}\cmu} ~~{Yash Parag Butala\textsuperscript{$\dagger$}\cmu} \\ {Melisa Russak}\writer~~ {Jing Yu Koh}\cmu ~~ {Kiran Kamble}\writer \\ {Waseem AlShikh}\writer ~~{Ruslan Salakhutdinov}\cmu \\ \cmu\textsl{Carnegie Mellon University} \writer\textsl{Writer.com} \\ \small\tt{\{raghavka, ypb\}@cs.cmu.edu}

\author{Raghav Kapoor\inst{1}* \and
Yash Parag Butala\inst{1}* \and
Melisa Russak\inst{2} \and
Jing Yu Koh\inst{1} \and
Kiran Kamble\inst{2} \and
Waseem AlShikh\inst{2} \and
Ruslan Salakhutdinov\inst{1}}

% %  FINAL: Replace with an abbreviated list of authors.
\authorrunning{R. Kapoor, Y. Butala et al.}
% % First names are abbreviated in the running head.
% % If there are more than two authors, 'et al.' is used.

% %  FINAL: Replace with your institution list.
\institute{Carnegie Mellon University \and Writer.com\\
\email{\{raghavka, ypb\}@cs.cmu.edu} \\
\url{https://huggingface.co/datasets/Writer/omniact}
}

\maketitle
% \twocolumn[{%
% \renewcommand\twocolumn[1][]{#1}%
% \maketitle
\begin{center}
    \centering
    \captionsetup{type=figure}
    \includegraphics[width=0.9\linewidth]{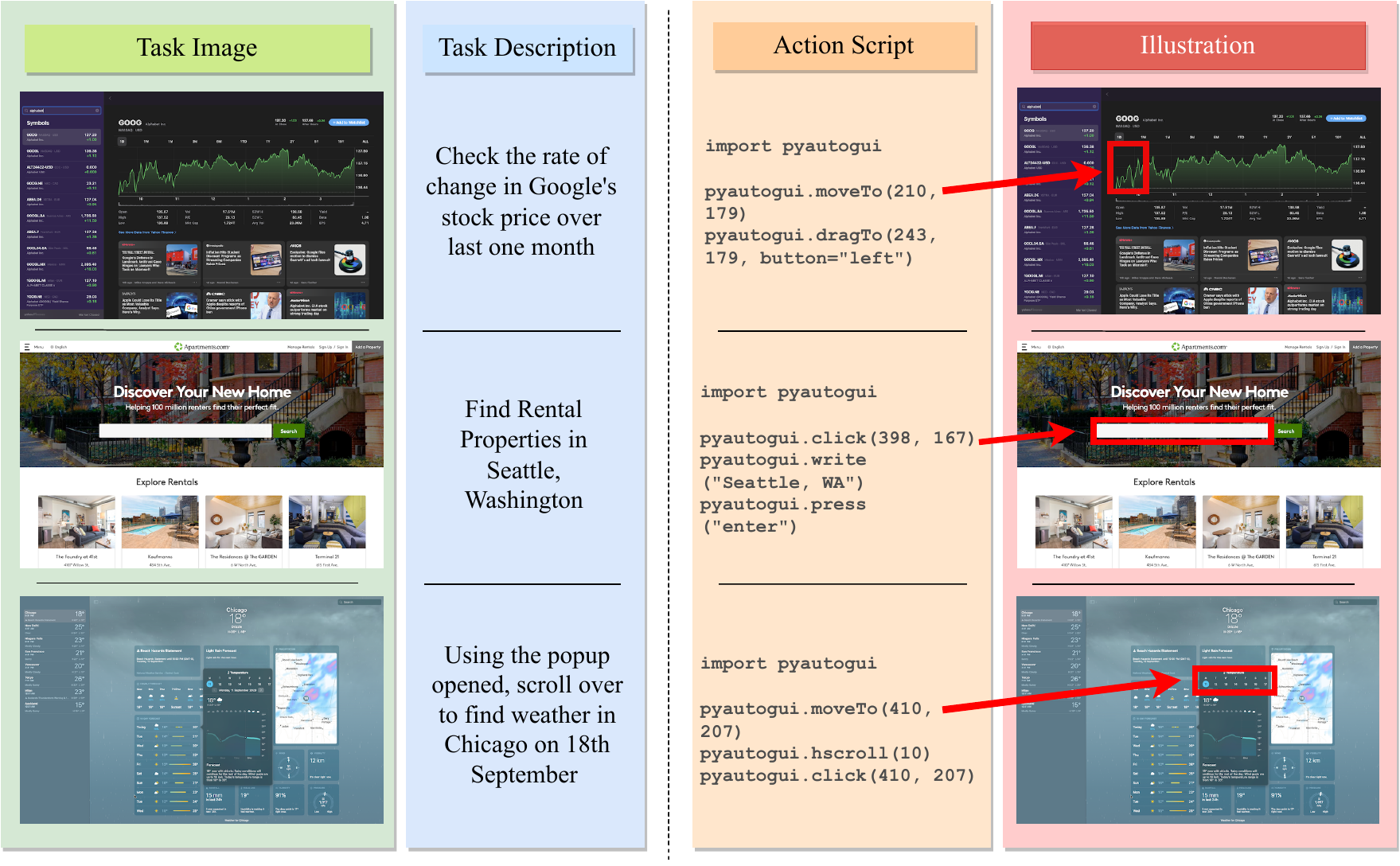}
    \captionof{figure}{\ModelName{} dataset and benchmark for enabling autonomous human-computer interaction agents. The left shows an image paired with a natural language task description as the input, the right shows the resulting action script to be executed on the screen. Examples are presented from Stocks, Apartments, and Weather application. }
    \label{fig:teaser}
\end{center}%
% }]

\begin{abstract}
\freefootnote{* These authors contributed equally. The order is determined by dice rolling.}
For decades, human-computer interaction has fundamentally been manual. Even today, almost all productive work done on the computer necessitates human input at every step. Autonomous virtual agents represent an exciting step in automating many of these menial tasks. Virtual agents would empower users with limited technical proficiency to harness the full possibilities of computer systems. They could also enable the efficient streamlining of numerous computer tasks, ranging from calendar management to complex travel bookings, with minimal human intervention. In this paper, we introduce \textit{\ModelName{}}, the first-of-a-kind dataset and benchmark for assessing an agent's capability to generate executable programs to accomplish computer tasks. Our scope extends beyond traditional web automation, covering a diverse range of desktop applications. The dataset consists of fundamental tasks such as ``Play the next song", as well as longer horizon tasks such as ``Send an email to John Doe mentioning the time and place to meet". Specifically, given a pair of screen image and a visually-grounded natural language task, the goal is to generate a script capable of fully executing the task. We run several strong baseline language model agents on our benchmark. The strongest baseline, GPT-4, performs the best on our benchmark However, its performance level still reaches only 15\% of the human proficiency in generating executable scripts capable of completing the task, demonstrating the challenge of our task for conventional web agents. Our benchmark provides a platform to measure and evaluate the progress of language model agents in automating computer tasks and motivates future work towards building multimodal models that bridge large language models and the visual grounding of computer screens.
\end{abstract}

\section{Introduction}
\label{sec:intro}

Performing computer tasks based on natural language instructions has been a long-standing goal of artificial intelligence \cite{wang2023survey}. One concrete objective in the line of research is to develop generalist agents that can assist humans in doing computer tasks \cite{lecun2022path}, such as \textit{``Order a pizza from Domino's"} or \textit{``Write a message to John."} The agent should be able to open the application and perform the task. Executing these actions on a personal computer involves a sequence of interactions with a mouse and keyboard. For example, the simple task of writing an email involves hovering over the application icon, clicking it, clicking the \textit{`New Email'} button, writing the content of the email, and clicking send. Successfully sending an email requires accurately predicting the correct action at each step and accurately executing it, which is a herculean task even for the best agents today \cite{gur2018learning}. % Libraries such as (Robotic Process Automation) RPA and \textit{\pyautogui} can execute kernel operations for controlling the mouse and keyboard.
% Such intelligent agents have a great potential to enhance the lives of the people with disabilities and those unable to access the technology. 

A generalist agent for computer tasks must understand natural language instructions, process visual screenshots, and produce the correct sequence of actions to be performed to achieve the intended task. %While it is possible to have agents that work with trial and error on the UI screens, there are challenges that come with that. These challenges can be related to 1) risks associated such as clicking on actions that can't be reversed, eg: purchasing a product. 2) the overhead time that arises due to wrong actions taken.  
Several existing approaches focus on building agents based on the HTML model~\cite{shi2017world,deng2023mind2web,zhou2023webarena}. However, this approach introduces several challenges and constraints. These agents are limited to web applications and often struggle with complex or long-context HTML code. They cannot interact with native desktop applications or perform tasks that span multiple applications, like drafting an email using text from a code editor, without significant alterations. Furthermore, HTML-based agents, which are inherently powered by text-only language models, typically underperform in tasks requiring visual cues, such as identifying and clicking a blue button on a desktop's top-right corner. In contrast, humans can easily understand UI elements like dropdown menus, typable areas, redirections, and options with just a glance.

% In order to build a truly generalist autonomous agent, strong UI understanding is a fundamental skill. There are works on `grounding referring expressions' or `functionality prediction' \cite{},  but they are far from building an agent that can convert natural language instructions to executable scripts. 

% Towards building truly general agents, capable of operating on web, inherently by means of multimodal skills, we propose a task and collect a dataset \ModelName\ consisting of 8k image instruction pairs (Figure~\ref{}). Given a screenshot of the current screen UI and natural language instructions describing the task, we want to generate the \textit{\pyautogui} code that achieves the task. We collected a set of tasks over diverse domains spanning both native applications and web browsers (Table~\ref{tab:dataset-2}). We collect tasks for native applications of three popular operating systems - MacOS, Windows, and Linux. 

Towards the goal of developing a generalist autonomous agent with robust visual and user interface (UI) understanding capabilities, we introduce a new task and dataset, \ModelName{}, containing over 9.8K pairs of images and instructions (Figure~\ref{fig:teaser}) across different operating systems and the web. This dataset includes screenshots of various UI screens and corresponding natural language instructions. The objective of these instructions is to generate executable commands using the \textit{\pyautogui} Python library~\cite{PyAutoGUI}. \textit{\pyautogui} enables the automation of the mouse and keyboard operations, which helps to facilitate interactions with various native applications across macOS, Windows, and Linux. This simplifies completing specified tasks across different web domains and native desktop applications.

We evaluate several language model-based agent baselines on this dataset, including LLaMA~\cite{touvron2023llama}, Vicuna~\cite{vicuna2023}, Palmyra-X (43B)~\cite{palmyra-x}, InstructPalmyra-30B~\cite{InstructPalmyra}, GPT 3.5, and GPT-4~\cite{openai2023gpt4}. We experiment with fine-tuning Vicuna-13B and LLaMA-13B models using QLoRA~\cite{dettmers2023qlora}. We also benchmark multimodal baseline LLaVa-v1.5-7B, LLaVa-v1.5-13B~\cite{touvron2023llama}, Gemini-Pro~\cite{team2023gemini} and GPT-4-vision-preview~\cite{yang2023dawn} for the task. Our findings highlight the necessity for a multimodal model capable of executing these tasks, and our analysis provides insights into promising future work in the space. 
Our key contributions are outlined as follows:
\begin{enumerate}
    \item We release a novel dataset of desktop and website applications consisting of over 9.8K natural language tasks, UI screens, and corresponding code snippets collected through human annotation. We introduce custom performance metrics tailored for computer tasks.
    \item We propose DetACT, a module for creating textual representations of the screen using signals from OCR, color, and icon-template matching.
    % \item We propose custom performance metrics tailored for computer tasks on desktop and web applications. %which take into account different actions and penalties. 
    \item We conduct a comprehensive benchmark and analysis of state-of-the-art LLMs and multimodal models on our benchmark. Our results show that \ModelName{} is a challenging task for even the best LLM agents today, and existing models are far below human performance.
\end{enumerate}

\section{Related Work}
\label{sec:related_work}

\begin{table*}[]
\centering
\caption{Comparison of OmniACT with other related benchmarks.}
\resizebox{\textwidth}{!}{%
\begin{tabular}{@{}ccccccccc@{}}
\toprule
\textbf{Datasets} &
  \textbf{Size} &
  \textbf{Env Type} &
  \textbf{\begin{tabular}[c]{@{}c@{}}Task \\ Heterogeneity\end{tabular}} &
  \textbf{\begin{tabular}[c]{@{}c@{}}Real-World \\ Portayal\end{tabular}} &
  \textbf{\begin{tabular}[c]{@{}c@{}}Executional \\ Correctness\end{tabular}} &
  \textbf{\begin{tabular}[c]{@{}c@{}}Supports \\ Desktop \\ Apps\end{tabular}} &
  \textbf{\begin{tabular}[c]{@{}c@{}}Continuous Scale \\ Adaptive \\ Evaluation\end{tabular}} &
  \textbf{Task} \\ \midrule
VisualWebArena~\cite{koh2024visualwebarena} &
  910 &
  Web &
  {\color[HTML]{32CB00} Yes} &
  {\color[HTML]{32CB00} Yes} &
  {\color[HTML]{32CB00} Yes} &
  {\color[HTML]{FE0000} No} &
  {\color[HTML]{FE0000} No} &
  Web Navigation \\
WebArena~\cite{zhou2023webarena} &
  812 &
  Web &
  {\color[HTML]{32CB00} Yes} &
  {\color[HTML]{32CB00} Yes} &
  {\color[HTML]{32CB00} Yes} &
  {\color[HTML]{FE0000} No} &
  {\color[HTML]{FE0000} No} &
  Web Navigation \\
Mind2Web~\cite{deng2023mind2web}&
  2350 &
  Web &
  {\color[HTML]{32CB00} Yes} &
  {\color[HTML]{32CB00} Yes} &
  {\color[HTML]{FE0000} No} &
  {\color[HTML]{FE0000} No} &
  {\color[HTML]{FE0000} No} &
  Web Navigation \\
WebShop~\cite{yao2022webshop} &
  12000 Products &
  Web &
  {\color[HTML]{FE0000} No} &
  {\color[HTML]{FE0000} No} &
  {\color[HTML]{32CB00} Yes} &
  {\color[HTML]{FE0000} No} &
  {\color[HTML]{FE0000} No} &
  Web Navigation \\
RUSS~\cite{xu2021grounding}&
  80 &
  Web &
  {\color[HTML]{32CB00} Yes} &
  {\color[HTML]{32CB00} Yes} &
  {\color[HTML]{FE0000} No} &
  {\color[HTML]{FE0000} No} &
  {\color[HTML]{FE0000} No} &
  Web Navigation \\
WebSRC~\cite{chen2021websrc} & 
  2735 &
  Web &
  {\color[HTML]{32CB00} Yes} &
  {\color[HTML]{32CB00} Yes} &
  - &
  {\color[HTML]{FE0000} No} &
  {\color[HTML]{FE0000} No} &
  QA \\ \midrule
MiniWoB++ \cite{humphreys2022data} &
  100 &
  \begin{tabular}[c]{@{}c@{}}Mobile \\ Websites\end{tabular} &
  {\color[HTML]{FE0000} No} &
  {\color[HTML]{FE0000} No} &
  {\color[HTML]{32CB00} Yes} &
  {\color[HTML]{FE0000} No} &
  {\color[HTML]{FE0000} No} &
  Web Navigation \\
PixelHelp~\cite{li2020mapping} &
  187 &
  Mobile &
  {\color[HTML]{32CB00} Yes} &
  {\color[HTML]{32CB00} Yes} &
  {\color[HTML]{FE0000} No} &
  {\color[HTML]{FE0000} No} &
  {\color[HTML]{FE0000} No} &
  UI Grounding \\
MetaGUI ~\cite{sun2022meta} &
  1125 &
  Mobile &
  {\color[HTML]{32CB00} Yes} &
  {\color[HTML]{32CB00} Yes} &
  {\color[HTML]{32CB00} Yes} &
  {\color[HTML]{FE0000} No} &
  {\color[HTML]{FE0000} No} &
  Mobile Navigation \\
MoTIF~\cite{burns2021mobile}&
  756 &
  Mobile &
  {\color[HTML]{32CB00} Yes} &
  {\color[HTML]{32CB00} Yes} &
  {\color[HTML]{32CB00} Yes} &
  {\color[HTML]{FE0000} No} &
  {\color[HTML]{FE0000} No} &
  Mobile Navigation \\
AITW~\cite{rawles2023android}&
  715142 &
  Mobile and Web &
  {\color[HTML]{32CB00} Yes} &
  {\color[HTML]{32CB00} Yes} &
  {\color[HTML]{32CB00} Yes} &
  {\color[HTML]{FE0000} No} &
  {\color[HTML]{FE0000} No} &
  \begin{tabular}[c]{@{}c@{}}Mobile/Web \\ Navigation\end{tabular} \\ %\toprule
  \midrule
\textbf{OmniACT} (Ours)&
  \textbf{9802} &
  \textbf{Desktop and Web} &
  {\color[HTML]{32CB00} \textbf{Yes}} &
  {\color[HTML]{32CB00} \textbf{Yes}} &
  {\color[HTML]{32CB00} \textbf{Yes}} &
  {\color[HTML]{32CB00} \textbf{Yes}} &
  {\color[HTML]{32CB00} \textbf{Yes}} &
  \textbf{Code Generation} \\ \bottomrule
\end{tabular}%
}

\label{tab:comparison}
\end{table*}

\subsection{UI Understanding}

User interface (UI) understanding has garnered interest from researchers in the machine learning and human-computer interaction communities, evolving with various models focusing on understanding the semantics of mobile and web user interfaces. UIBert~\cite{bai2021uibert}, PixelBERT~\cite{huang2020pixelbert}, ActionBert~\cite{he2021actionbert}, VUT~\cite{li2021vut}, Screen2Words~\cite{wang2021screen2words}, WidgetCaptioning~\cite{li2020widget} and Pix2Act~\cite{shaw2023pixels} are  notable models in this area. They propose approaches for learning the user-interface semantics of the mobile screen using the image and view hierarchy. These models have demonstrated effectiveness in tasks like capability prediction, screen segmentation and understanding, and screen caption generation. Lexi~\cite{banerjee2023lexi} and Spotlight~\cite{li2023spotlight} propose models that use vision-only inputs to minimize the reliance on metadata such as view hierarchy. Furata et al. \cite{furuta2023multimodal} demonstrates the use of fine-tuning for multimodal web navigation. The majority of machine learning models trained for UI understanding leverage the Rico dataset~\cite{rico} and its extensions, which contain 64,462 unique Android screens and metadata. In addition, \cite{banerjee2023lexi} released the UICaptions dataset, which consists of diverse image-captions pairs across a wide range of applications. PixelHelp~\cite{li2020mapping} also released a corpus to train models that can interpret natural language instructions and map them to mobile UI actions.

\subsection{Autonomous Computer Agents}
% The onset of large language models (LLMs) has led to rapid growth in improvising agents that operate on web pages. Many of the recent research works, such as ViperGPT~\cite{surís2023vipergpt} and Chameleon~\cite{lu2023chameleon}, leverage large language models for planning or action prediction, to build autonomous agents. Further, datasets like  Mind2Web \cite{deng2023mind2web} and WebArena \cite{zhou2023webarena} aim to automate the web by using text-based approaches. However, this limits the ability of autonomous agents only to the web as these agents operate on a text-based DOM model of the HTML script of a webpage and do not understand the screen context, which serves as an important signal for the model to take action. \cite{rawles2023android} release a dataset Android in the Wild consisting of screens, natural language instructions, and corresponding actions. \cite{autoui} proposes a multimodal model AutoUI to build an agent on the \cite{rawles2023android} dataset. However, they are restricted to the Android environment. Recently, there have been attempts, such as Auto-GPT \footnote{\href{https://github.com/Significant-Gravitas/AutoGPT}{https://github.com/Significant-Gravitas/AutoGPT}}, to build an end-to-end pipeline involving text-only LLMs. 
The advent of large language models (LLMs) has been pivotal in the rapid advancement of agents that operate on web pages. Recent research such as ViperGPT~\cite{surís2023vipergpt} Chameleon~\cite{lu2023chameleon}, RCI Agent~\cite{kim2023language}, VisProg~\cite{gupta2023visual}, and \cite{nakano2021webgpt} employ LLMs for planning or action prediction in developing autonomous agents. Benchmark datasets, such as MiniWoB~\cite{shi2017world}, WebShop~\cite{yao2022webshop}, 
Macaw-LLM~\cite{lyu2023macaw},
ASH-Prompting~\cite{sridhar2023hierarchical}
Mind2Web~\cite{deng2023mind2web}, WebArena~\cite{zhou2023webarena}, AgentBench~\cite{liu2023agentbench}  and VisualWebArena \cite{koh2024visualwebarena} 
have also been proposed to measure the ability of LLM-based agents to automate web tasks. These methods mainly involve agents that operate on a text-based Document Object Model (DOM) of HTML scripts. This limits their understanding of screen context, which is crucial for the model's decision-making and action-taking processes. To address this limitation, \cite{rawles2023android} released Android in the Wild, a dataset comprising screens, natural language instructions, and corresponding actions. Following this, ~\cite{autoui} proposed a multimodal model, AutoUI, which is designed to build an agent on the Android in the Wild dataset confined to the Android ecosystem. WebAgent \cite{gur2024realworld-WebAgent} utilized Flan-U-PaLM, for grounded code generation, and HTML-T5 and showed improvement on real-world websites.

Current benchmarks for autonomous agents focus mainly on the Web or Android environments, posing challenges for tasks involving desktop applications or spanning multiple applications beyond the web domain. The absence of established benchmarks and datasets in this area, coupled with basic methods for extracting user interface (UI) elements, underscores the need for significant progress in developing more versatile autonomous agents capable of handling diverse tasks beyond the current scope. To highlight the unique features that \ModelName{} introduces in the assessment of capable autonomous agents, we provide a comparison between the existing benchmarks and our proposed benchmark, \ModelName{}, in Table \ref{tab:comparison}.

\begin{figure*}[!ht]
    \centering
    \includegraphics[width=0.9\linewidth]{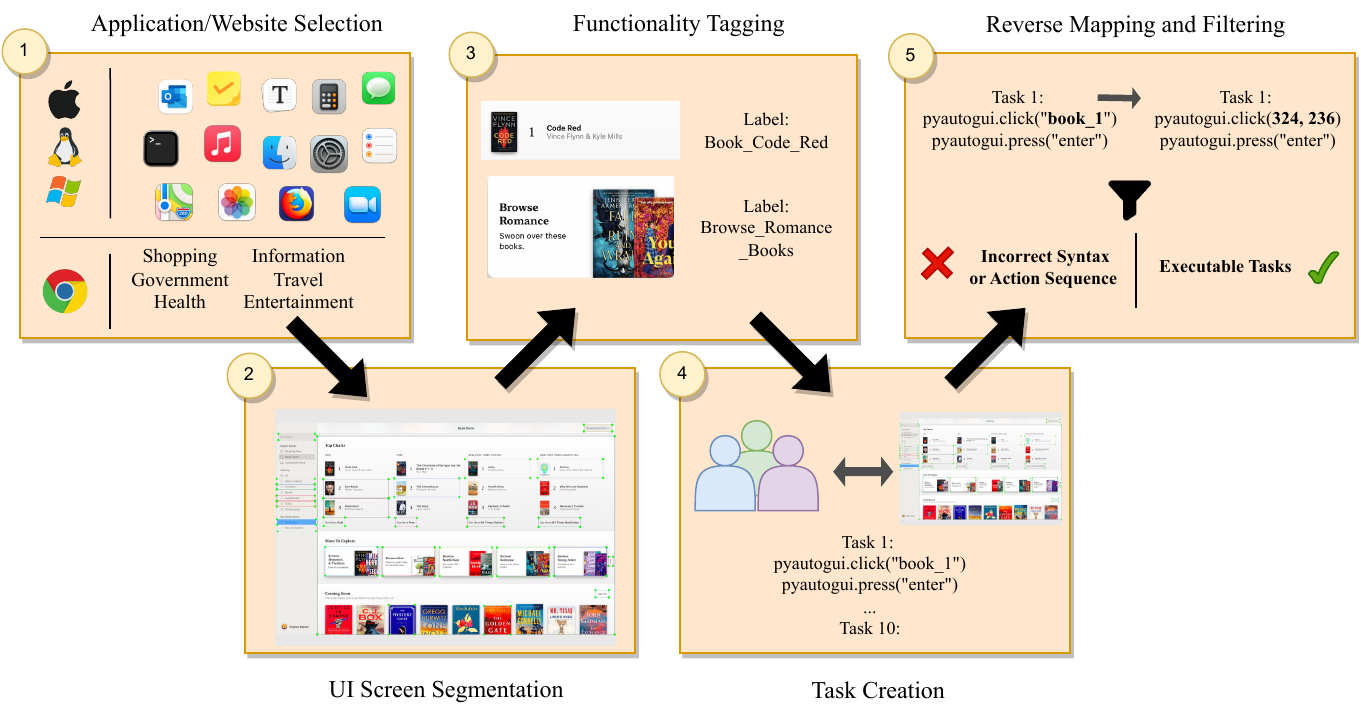}
    \caption{\textbf{Data Collection Pipeline.} (1) We select over 60 applications and websites to ensure diversity, (2) segment the screen through human-annotated bounding boxes, (3) label the bounding boxes based on functionality, (4) ask student volunteers to come up with tasks, given a screen image, and (5) reverse map the textual labels to coordinates and filter the scripts based on execution and syntax.}
    \label{fig:dataset}
\end{figure*}
\section{\ModelName{}}

We introduce a novel dataset and benchmark, \ModelName{}, which measures the performance of autonomous agents on both web and desktop applications. Compared to previous benchmarks which focus on text-based reasoning~\cite{shi2017world,zhou2023webarena,deng2023mind2web, yao2022webshop, humphreys2022data}, our benchmark aims to measure multimodal agents that bridge large language model planners and UI understanding vision models. \ModelName{} can be accomplished as a standalone task as it is not under a mock environment. 

All actions that a human can execute on the computer can be encoded in the \textit{\pyautogui}~\cite{PyAutoGUI} Python framework. This framework allows a user to execute keyboard and mouse operations by running Python code. The \textit{\pyautogui} code to execute these tasks is shown in the third column of Figure~\ref{fig:teaser}. For other computer tasks, the \textit{\pyautogui} library provides functions such as `press', `write', and `scroll' which can be used to execute the task. Our dataset consists of parallel data of natural language tasks, UI screenshots, and ground truth \textit{\pyautogui} scripts that achieve successful execution.

\subsection{Task Formulation}

Given an input state of a computer defined by the screen $S$ and the task description $T$ in natural language, the goal of the task is to output a sequence of actions $A$ that can successfully accomplish the task $T$ within a screenshot $S$ $\in \{\text{Linux, Windows, MacOS, Webpage}\}$. Formally, the task can be defined as learning the transition function $f: T \times S \rightarrow A$. During dataset collection, we ensure that all task descriptions $T$ are feasible and can be accomplished in the current screenshot $S$. To reduce ambiguity and facilitate better evaluation, we ensure that task descriptions are detailed and unambiguous. Tasks can also be visually grounded (e.g., `Click the red button to start recording') or natural language based (e.g., `Click the My Account button'). We define the action space using the functionalities in the \textit{\pyautogui} library: $A \in \{\text{`click', `dragTo', `scroll', `write'}, \ \ldots \}$. The exhaustive list of actions is provided in Table~\ref{tab:action}. Our action space is much larger than other benchmarks ~\cite{shi2017world,deng2023mind2web,zhou2023webarena} that resort to two or three interaction options. Mouse actions such as `moveTo', `click', `rightClick', `doubleClick', and `dragTo', additionally require screen coordinates as arguments, which indicate the pixel location of the action. 

Figure~\ref{fig:teaser} illustrates sample tasks and corresponding outputs for three applications within \ModelName: (1) Stocks (MacOS), (2) Apartments.com (web page), and (3) Weather (MacOS). The first column depicts the input image, and the second column shows the natural language task that is to be executed on the current screen. To execute these tasks, a user must accurately perform a series of operations using the mouse and keyboard. Eg: to check the rate of change in Google's stock price over the last month, the mouse has to be moved to the last month and dragged while holding the left-click button to the current month.

\subsection{Dataset Preparation}
% We select applications and websites, create bounding boxes, and label them based on their functionality so that it is easier for the volunteers to write the script for the task. Later, we map the labels back to numeric coordinates and filter the scripts based on execution and syntax. To curate our dataset, we use a pipelined approach (Figure~\ref{fig:dataset}), which we describe in detail in the following sections.
To prepare our dataset, we followed a pipelined approach, as summarized in Figure~\ref{fig:dataset}. We first selected a variety of applications and websites. For each application or website, we created bounding boxes around key UI elements and labeled them according to their functionality, which is crucial for assisting human annotators in writing accurate \textit{\pyautogui} scripts. After each script is written, we converted the labels back into numeric coordinates, allowing us to align the scripts precisely with the locations of the UI elements. Finally, we thoroughly reviewed each script, focusing on its executability and adherence to syntax standards. This ensured the high quality and functionality of our dataset, making it a valuable resource for training and evaluating autonomous agents.

% Please add the following required packages to your document preamble:
% \usepackage{booktabs}
% \usepackage{multirow}
% \usepackage{graphicx}
% \usepackage[table,xcdraw]{xcolor}
% Beamer presentation requires \usepackage{colortbl} instead of \usepackage[table,xcdraw]{xcolor}
\begin{wraptable}{R}{5.5cm}
\centering
\caption{Action types supported by \ModelName\ and the number of instances for each action in the dataset.}
\resizebox{0.25\textwidth}{!}{%
\begin{tabular}{@{}ccr@{}}
\toprule
\multicolumn{1}{l}{\textbf{Type}}                                               & \textbf{Action}            & \multicolumn{1}{c}{\textbf{\%}} \\ \midrule
\rowcolor[HTML]{FFFFC7} 
\multicolumn{1}{c}{} & Click             & \multicolumn{1}{c}{63.73}                  \\
\rowcolor[HTML]{FFFFC7} 
\multicolumn{1}{c}{} & Double Click      & \multicolumn{1}{c}{0.58}                   \\
\rowcolor[HTML]{FFFFC7} 
\multicolumn{1}{c}{} & Right Click       & \multicolumn{1}{c}{0.77}                   \\
\rowcolor[HTML]{FFFFC7} 
\multicolumn{1}{c}{} & Move/Hover        & \multicolumn{1}{c}{1.85}                   \\
\rowcolor[HTML]{FFFFC7} 
\multicolumn{1}{c}{} & Drag              & \multicolumn{1}{c}{0.29}                   \\
\rowcolor[HTML]{FFFFC7} 
\multicolumn{1}{c}{} & Scroll            & \multicolumn{1}{c}{1.68}                   \\
\rowcolor[HTML]{FFFFC7} 
\multicolumn{1}{l}{\multirow{-7}{*}{\textbf{Mouse}}}    & Horizontal Scroll & \multicolumn{1}{c}{0.17}                   \\ \midrule
\rowcolor[HTML]{C3EFAF} 
\multicolumn{1}{c}{} & Press             & \multicolumn{1}{c}{16.28}                  \\
\rowcolor[HTML]{C3EFAF} 
\multicolumn{1}{c}{} & Hotkey            & \multicolumn{1}{c}{3.00}                      \\
\rowcolor[HTML]{C3EFAF} 
\multicolumn{1}{l}{\multirow{-3}{*}{\textbf{Keyboard}}}  & Write             & \multicolumn{1}{c}{11.65}                  \\ \bottomrule
\end{tabular}%
}

\label{tab:action}
\end{wraptable}

\subsubsection{Application/Website Selection}
To test the computer agents' generalization ability across different tasks, we collect tasks across multiple domains on both desktop and web applications. In total, we collect and annotate 9802 data points (Table~\ref{tab:dataset-2}), with the split between desktop and web applications approximately 3:1. The emphasis on desktop applications, which do not contain Document Object Model (DOM) hierarchies unlike HTML-based web pages, presents a more complex multimodal challenge where visual cues are crucial. We collect tasks from applications within the three most popular operating systems. We select 22 native applications from MacOS, and 8 each from Linux and Windows. We annotate roughly 3 to 4 screens for every application. The full list of applications is provided in the Appendix. 

Many common computer tasks today are still performed through web applications, so we also collect 3-4 screenshots from 27 different web applications. To ensure diversity in task intents, we categorize these tasks into one of the following 6 categories: (1) Shopping, (2) Entertainment, (3) Service, (4) Government, (5) Travel, (6) Health. Inspired by the methodology of~\cite{deng2023mind2web}, these categories were selected to cover a wide range of user intents and functionalities.

\subsubsection{UI Screen Segmentation}
To collect gold-standard data, we first annotate and segment the screen by identifying the bounding boxes present on the screen. We employ slightly different techniques for web and desktop applications to create the bounding boxes:

\begin{enumerate}
    \item \textbf{Desktop Applications:} We build a custom annotation interface based on PyQt5\footnote{\url{https://pypi.org/project/PyQt5/}} to create bounding boxes manually over a screen image using a simple drag-and-click mechanism. This custom interface expedites the process and allows us to get highly accurate gold-label data for desktop images.
    \item \textbf{Websites:} For webpages, we write JavaScript code to extract all interactable (click,  type, etc.) regions from HTML source code. We also extract banners, dropdowns, submit, and radio buttons from the screen. We filter the elements to retain only those that are visible and interactable within the screen.
\end{enumerate}

\subsubsection{Functionality Tagging}

To map each bounding box to its correct functional description, we leverage Amazon MTurk workers (see details in Appendix), who are given an image with a bounding box and are required to write the correct description or label of the bounding box's function. For example, given an image of an Amazon webpage with a \textit{search bar}, the annotator labels it as \textit{``find-product-search-bar"}. The logical descriptions are used to create tasks in a structured manner without the need to identify individual bounding box coordinates.

\subsubsection{Task Creation}

\begin{figure*}[!ht]
    \centering
    \includegraphics[width=\linewidth]{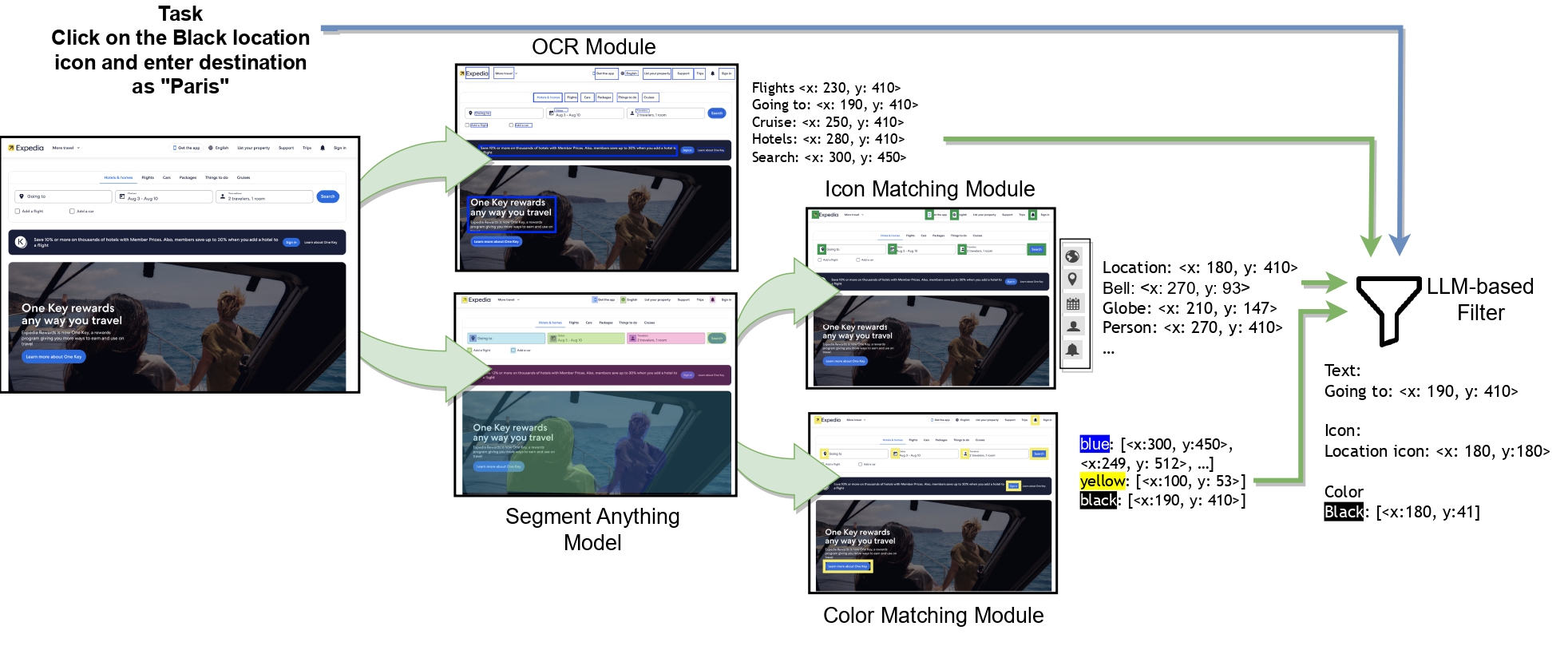}
    \caption{\textbf{DetACT Module.} Given an initial image and a natural language task description, we use a pipelined approach to run OCR and SAM on the screen. The outputs from SAM are then used by icon and color-matching modules to obtain an exhaustive set of useful UI elements. The list of elements is passed through LLM based filter to select only the elements related to the given task.}
    \label{fig:detact}
\end{figure*}

% For every screen, we aim to leverage all the human-annotated bounding boxes and corresponding labels to form tasks that can be completed without exiting the screen. As the tasks are aimed at improving the fundamental understanding of the agent, we ensure that the tasks more visually grounded.

Our approach for each screen involves utilizing all human-annotated bounding boxes and their labels to create tasks that can be executed within the confines of a single screen. These tasks are designed to be visually grounded in order to measure the capabilities of multimodal agents. We plan to release the bounding box and their corresponding labels as the metadata for evaluation purposes.

% We recruit college student annotators with basic programming knowledge in Python, and provide API references for \textit{\pyautogui} and example tasks. Each student is tasked to devise multiple tasks along with three possible reformulations of the natural language description of the task. \textbf{We ensure that rephrased tasks fall into in only one of the splits to avoid overfitting.} We provide further details about this process in Appendix~\textbf{??}.

For dataset compilation, college students with basic Python programming skills served as annotators, accessing API references for \textit{\pyautogui} and examples of potential tasks. Each student generated multiple tasks, each accompanied by three alternative natural language reformulations. For instance, \textit{``What is 3+2?"} might be reformulated as \textit{``Calculate the sum of 2 and 3"} or \textit{``Add two to three"}. To avoid train-test leakage, rephrased tasks were consistently placed in the same dataset split. Further details on the annotation process are available in the Appendix.

\subsubsection{Reverse Mapping and Filtering}

\begin{wraptable}{R}{5.5cm}
\centering
\caption{Dataset distribution across splits and platforms.}
\resizebox{0.4\columnwidth}{!}{%
\begin{tabular}{@{}cccccc@{}}
\toprule
\multicolumn{2}{c}{\textbf{Domain}}                                         & \textbf{Train} & \textbf{Validation} & \textbf{Test}  & \multicolumn{1}{c}{\textbf{Total}} \\ \midrule
\rowcolor[HTML]{FFFFC7} 
\multicolumn{1}{c}{\cellcolor[HTML]{FFFFC7}} & Mac OS               & 3028  & 444        & 786   & \multicolumn{1}{c}{\cellcolor[HTML]{FFFFC7}4258}  \\
\rowcolor[HTML]{FFFFC7} 
\multicolumn{1}{c}{\cellcolor[HTML]{FFFFC7}} & Linux                & 761   & 126        & 247   & \multicolumn{1}{c}{\cellcolor[HTML]{FFFFC7}1134}  \\
\rowcolor[HTML]{FFFFC7} 
\multicolumn{1}{c}{\multirow{-3}{*}{\cellcolor[HTML]{FFFFC7}Desktop}} & Windows & 1573 & 216 & 458 & \multicolumn{1}{c}{\cellcolor[HTML]{FFFFC7} 2247} \\
\rowcolor[HTML]{C3EFAF} 
\multicolumn{1}{c}{\cellcolor[HTML]{C3EFAF}Web}                                           &          -            & 1427  & 206        & 530   & \multicolumn{1}{c}{\cellcolor[HTML]{C3EFAF} 2163}  \\ \midrule
\textbf{Total}                                         &                      & 6789  & 992        & 2,021 & 9802  \\ \bottomrule
\end{tabular}%
}
\label{tab:dataset-2}
\end{wraptable}

To ensure high-quality data, we incorporate an additional step into the data collection pipeline. We build scripts to map the text-based labels of each bounding box back to their numeric coordinates, and then match the syntax and verify if the task will be executed on the screen. Using this filter, we remove all the non-working or syntactically incorrect data points and finally manually review the set of tasks.

After filtering, we obtain 9802 human-annotated, gold-label data points across more than 200 desktop and web screens (Table~\ref{tab:dataset-2}), split into train, validation, and test sets in a 7:1:2 ratio. All collected data will be publicly released to encourage future work on multimodal agents.

\section{Evaluation Metrics}

In this section, we detail various evaluation metrics for benchmarking model performance on the \ModelName\ dataset. UI screens have additional constraints such as spatial relevance which are not factored in most conventional similarity-based metrics such as BLEU~\cite{papineni2002bleu}, CodeBLEU~\cite{ren2020codebleu}, BERTScore \cite{zhang2020bertscore} and CodeBERTScore~\cite{zhou2023codebertscore}. For example, a valid click action is usually not constrained to a single coordinate but can be any coordinate within a specified region. In the event of invalid coordinate predictions, an agent that predicts coordinates further away from the valid region should invoke a higher penalty compared to an agent that predicted coordinates close to the region. We propose two new metrics adapted: Sequence Score (Section~\ref{sec:sequence_score}) and Action Score (Section~\ref{sec:action_score}) aimed at utilizing UI information. 

\subsection{Sequence Score}  \label{sec:sequence_score}
The sequence score measures whether the predicted action sequence (e.g., `click', `write', `press') exactly matches the gold sequence. Since predicting the first action in the sequence is relatively straightforward and later actions are more difficult, we define sequence score as follows: \setlength{\abovedisplayskip}{3pt}
\setlength{\belowdisplayskip}{3pt}
\begin{equation}
SeqScore_i = 
    \begin{cases}
        \beta_1 + \beta_2 * (s-1) & \text{if all actions match} \\
        0 & \text{otherwise}
    \end{cases}
\end{equation}
where $s$ is the action sequence length, $\beta_1$ is set to 0.1 and $\beta_2$ is set to 1.
\subsection{Action Score}  \label{sec:action_score}
The action score measures how well a code snippet containing the correct action sequence can perform the task. Specifically, for a script with a correct action sequence, we introduce penalties for inaccurate behavior. The penalties are described below:

\begin{enumerate}
    \item \textbf{Click penalty ($M$)}: For actions  `click', `rightClick', `doubleClick', `moveTo', and `dragTo', we penalize code snippets where predicted coordinates lie outside of the bounding box of the UI element. The click penalty for the $j^{th}$ action of the $i^{th}$ example is defined as:
\begin{equation}
\textbf{$M_i^j$} =  \alpha_i \times 
\begin{dcases}
    1 - \frac{\mu}{\mu+L_2} & \text{if $SeqScore_i$} > 0\\
    1 & \text{otherwise}
\end{dcases}
\end{equation}

Here $L_2$ corresponds to the smallest Euclidean distance between the predicted coordinate and bounding box. $ L_2$ is zero when the predicted coordinate lies within the target bounding box. $\mu$ is the Dirichlet smoothing coefficient which we dynamically set to the inverse of the length of the diagonal of the bounding box. This ensures that the penalty for points outside the bounding box varies based on the size of the bounding box. For two predicted points with the same $L_2$, the metric penalizes more heavily if the box is larger. This is sound with the intuition that the chances of clicking on a larger box are higher and should be penalized more in case of a mistake. 
    \item \textbf{Key penalty ($K$)}: For actions `press' and `hotkey', we check whether the set of keys in the target code (represented as $GK_i^j$) and predicted code (represented as $PK_i^j$) are the same. It is formally defined as:
\setlength{\abovedisplayskip}{3pt}
\setlength{\belowdisplayskip}{3pt}
\begin{equation}
\textbf{$K_i^j$ = }
\alpha_i \times
  \begin{cases}
    0 & \text{if  }GK_i^j = PK_i^j \text{ and } SeqScore_i > 0\\
    1 & \text{otherwise}
  \end{cases}
\end{equation}
    \item \textbf{Write penalty ($W_p$)}: For action type `write', we penalize the output for the sentence to be typed. Specifically, we the employ BLEU score~\cite{papineni2002bleu}, and compute:
\setlength{\abovedisplayskip}{3pt}
\setlength{\belowdisplayskip}{3pt}
\begin{equation}
\textbf{$W_i^j$ = }
\alpha_i \times
\begin{cases}
    \text 1- {BLEU}(GS_i^j, PS_i^j) & \text{if } SeqScore_i>1 \\
    1 & \text{otherwise}    
\end{cases} 
\end{equation}

Here, $GS_i^j$ represents the actual sentence to be typed, and $PS_i^j$ represents the sentence predicted by the model in the $j^{th}$ action of example $i$. 
\end{enumerate}
In the above equations, ($\alpha_i$) is the weighting factor:
\setlength{\abovedisplayskip}{3pt}
\setlength{\belowdisplayskip}{3pt}
\begin{equation}
    \alpha_i = SeqScore_i / \text{ length of sequence $i$}
\end{equation}

This ensures that the action score $\in [0,1]$. The mean action score is calculated as follows:
\setlength{\abovedisplayskip}{3pt}
\setlength{\belowdisplayskip}{3pt}
\begin{equation}
\textbf{Action Score = } 
\frac{ \sum_{i} max  \left (SeqScore_i -  \sum_j  (M_i^j + K_i^j+ W_i^j) , 0\right )}{\sum_{i} SeqScore_i}
\end{equation}

\section{ DetACT: DETecting ACTions from UI}

\begin{figure}[!ht]
    \centering
    \includegraphics[width=0.6\linewidth]{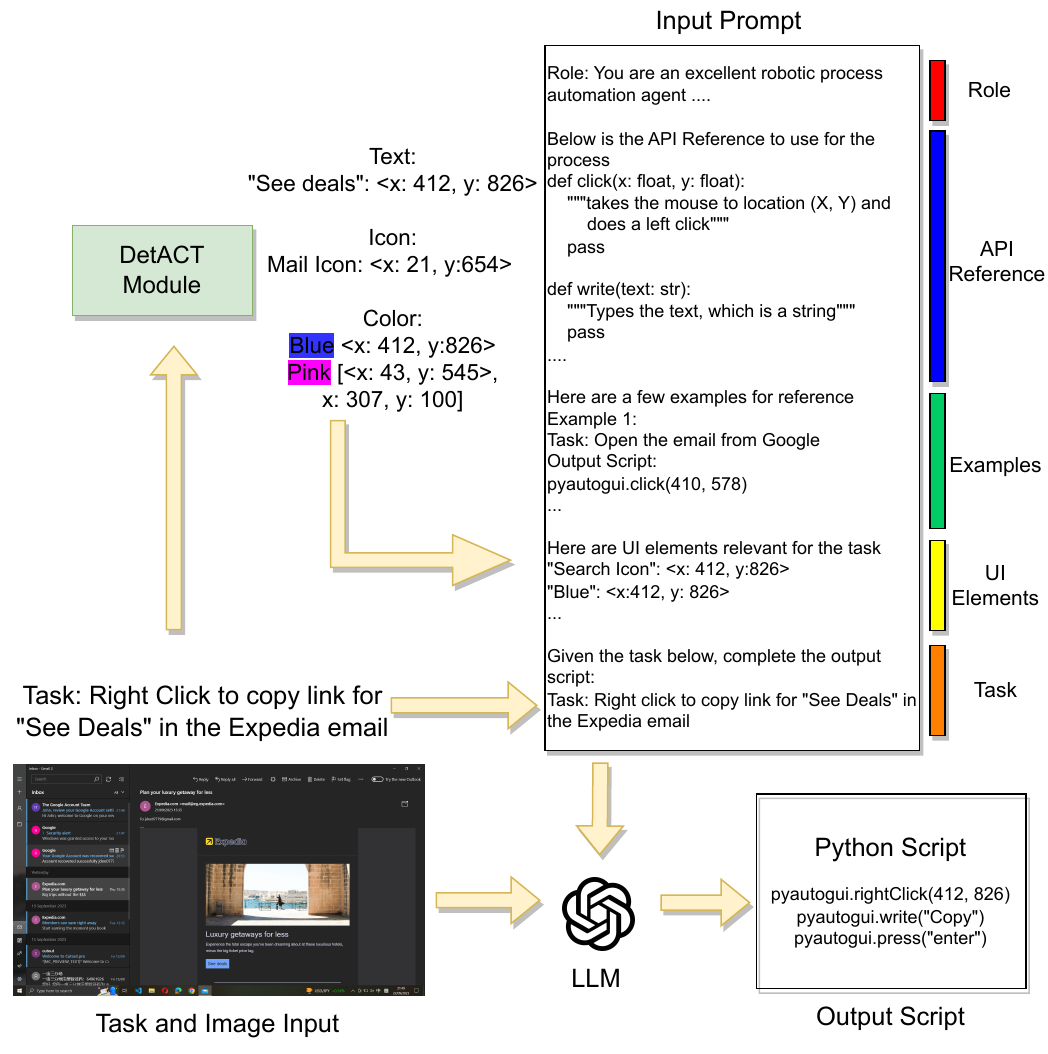}

    \caption{\textbf{Baseline Model Architecture.} Image and task descriptions are sent to DetACT module, which gives a filtered list of UI elements relevant to feed into the prompt along with the task. We also show the prompt structure used for action script generation. This structure is passed through the LLM (along with the image for multimodal LLM) to generate the automation script.}

    \label{fig:model}
    
\end{figure}

Understanding UI screens is crucial for multimodal computer tasks. Web-based agents typically use language-only inputs from the HTML DOM. This is insufficient for comprehending the full extent of an application UI, as many components may not be easily described with HTML code. To address this, we propose DetACT, which allows us to convert images of UI layouts into structured code and text outputs for a downstream LLM. DetACT is a system comprised of three distinct modules: the text module, the icon module, and the color module.

\begin{enumerate}
    \item \textbf{Text Extraction:} We use the EasyOCR model\footnote{\url{https://github.com/JaidedAI/EasyOCR}} to parse over the UI screens and collect all text-based elements. Along with the text, we also note the locations of each of these elements. This is depicted in Figure \ref{fig:detact}, along with a list of text elements found on the screen using the OCR Module. We segment and classify the different regions within the screenshot using the Segment Anything Model (SAM)~\cite{kirillov2023segment}. From the outputs, we filter out the non-textual segments for our icon and color detection.%, as described below.
    
\item \textbf{Icon Module:} For matching with the appropriate icon, we use a pack of 1600 icons\footnote{\url{https://icomoon.io/}} as templates. Each of these icons is labeled with their appropriate functionality and is matched with the filtered outputs SAM~\cite{kirillov2023segment}. For the similarity of the two images, we resize the reference icons and segmented region of interest (ROI) to the same size, and convert both images to grayscale. After this, we use the Structural Similarity Index (SSIM)~\cite{wang2004image}, to find the closest match of the ROI to the icons in our set, and select the ones above the SSIM threshold of 0.95. As seen in Figure~\ref{fig:detact}, a few icons matched on the screen are \textit{Globe} icon, \textit{Calendar} icon, \textit{Person} icon, and \textit{Location} icon; each depicting a different use case. % for the travel application.
\item \textbf{Color Module:} Finally, to place all segments of interest into appropriate buckets of colors, we average the RGB pixel values over the ROI and, based on that value, bucket them into different color categories. We categorize colors differently based on the human perspective of the ranges of each color. To avoid ambiguity, we consider eleven major colors, namely yellow, blue, green, red, pink, violet, white, black, orange, brown, and grey. We record the center of the element along with the color. 
\end{enumerate}

Once all the elements of each category are extracted with their coordinates, we then filter these UI elements by prompting GPT-4 \cite{openai2023gpt4}. We ensure that the elements selected are suited only for our task, for which we also provide the task description in our prompts along with the list of elements. Full details of the prompt are provided in the appendix section of the paper. As we observe in Figure~\ref{fig:detact}, given an image from the Expedia application, and a task (\textit{``Click on the Black Location icon and enter the destination as Paris."}), the LLM filters out the elements to retain only \textit{``Going To"}, \textit{``Location Icon"}, and the \textit{Black} colored elements from the screen. This is passed as input to the LLM or VLM backbone. %, which we discuss in the following section. 

\section{Baselines}

To evaluate the performance of existing language model-based agents, we conduct experiments with both language-based and multimodal baselines. The DetACT module takes in image and text descriptions of the task and outputs the color, icon, and text-based signals. This is concatenated to the prompt for the LLM prompt-based baselines (see Figure~\ref{fig:model}). Every prompt starts with a role assignment~\cite{zhao2023survey}, followed by the detailed API reference of the \textit{\pyautogui} function set, along with a textual description of their function. We then add five in-context examples from the training set that most closely match the task (based on the cosine similarity of the MiniLM~\cite{wang2020minilm} embeddings of the reference task and the train examples). We add a list of UI elements filtered by the DetACT module to the prompt. Finally, we provide the rules with the task description. For multimodal baselines, we also pass the image pixels to the vision encoder. We choose coordinate-based UI elements in the prompt as recent techniques like the Set-of-Mark (SOM)~\cite{yang2023setofmark} prompting does not work for desktop settings since it is difficult to obtain interactive elements from the desktop screen images. We report the results of several baselines:
\begin{itemize}
    \item \textbf{Few-shot Generative LLM:} 
    We experiment with models from LLaMA-2~\cite{touvron2023llama}, Vicuna-1.5~\cite{vicuna2023}, CodeLLaMA-34B~\cite{rozière2023code}, Palmyra~\cite{Palmyra}, and GPT \cite{openai2023gpt4} series. We use the prompts structure as shown in Figure~\ref{fig:model} to prompt the model. For LLaMA and CodeLLaMa, we reduce the prompt length to 2000 tokens by removing outputs from the DetACT module with lower confidence, as we observed poor performance on longer prompts. For the other models, we allow prompts with up to 4000 token sizes.
    
    \item \textbf{Finetuned Generative LLM:} 
    We fine-tuned the LLaMA-13B model and Vicuna-13B using QLoRa~\cite{dettmers2023qlora} with rank 64 and scaling factor 16 for 300 steps to generate the code given screen description from the DetACT module and the instruction.% We detail the hyperparameters and training details in the Appendix.

    \item \textbf{Few-shot Generative Multimodal Models:}
    As \ModelName\ is predominantly multimodal, with a majority of tasks being visually grounded, we conduct experiments with large multimodal models. Given the limited research in this domain \cite{mlm_survey, mlm_survey2}, there is a scarcity of available multimodal models with significant size adept for this task. Here, we experiment with \cite{liu2023llava, liu2023improvedllava}, providing a similar prompt as well as the screen image.% in addition to text. 
\end{itemize}

\section{Results and Analysis}
\label{sec:results}

\begin{wraptable}{R}{5.5cm}
\centering
\caption{Baseline Performance. (A) Prompt-only LLMs, (B) Fine Tuned LLMs, (C) Prompt-only Multimodal Models. The table represents the Sequence score (SS), click penalty ($M_p$), Key penalty ($K_p$), Write Penalty ($W_p$), and Action Score (AS). The best results for the (SS) and (AS) are highlighted.}
\resizebox{0.45\columnwidth}{!}{%
\begin{tabular}{@{}lcccccc@{}}
\toprule
\textbf{Model} &
  \textbf{\begin{tabular}[c]{@{}c@{}}SS($\uparrow$)\end{tabular}} &
  \textbf{\begin{tabular}[c]{@{}c@{}}$M_p$\end{tabular}} &
  \textbf{\begin{tabular}[c]{@{}c@{}}$K_p$\end{tabular}} &
  \textbf{\begin{tabular}[c]{@{}c@{}}$W_p$\end{tabular}} &
  \textbf{\begin{tabular}[c]{@{}c@{}}AS($\uparrow$)\end{tabular}}  \\ \midrule
\textbf{\small Prompt based LLMs} \\
LLaMA-7B~\cite{touvron2023llama}              & 4.12  & 1.24  & 1.83 & 0.57 & 0.48  \\
Vicuna-7B ~\cite{vicuna2023}            & 3.88  & 1.17  & 1.51 & 0.43 & 0.77  \\
LLaMA-13B ~\cite{touvron2023llama}            & 4.80   & 1.32  & 0.93 & 0.93 & 1.62\\
Vicuna-13B ~\cite{vicuna2023}           & 5.44   & 1.65  & 0.94  & 1.06 & 1.78  \\
Palmyra-Instruct-30B \cite{InstructPalmyra} & 7.51   & 5.68  & 0.12 & 0.40  & 1.31  \\
CodeLLaMA-34B \cite{rozière2023codellama}        & 10.09 & 2.99  & 2.71 & 0.66 & 3.72 \\
Palmyra-X 43B \cite{palmyra-x}        & 11.20  & 3.12  & 3.02 & 2.12 & 2.94 \\
GPT-3.5-turbo-0613~\cite{chatgpt-citation}       & 22.85 & 8.13  & 4.51 & 2.31  & 7.89 \\
GPT-4  \cite{openai2023gpt4}               & \textbf{32.75} & 10.27 & 6.99 & 3.89 & \textbf{11.60} \\ \midrule
\textbf{\small Finetuned LLMs} \\
LLaMA-13B FT          & \textbf{8.92}  & 4.61  & 1.43 & 0.74 & 2.14  \\
Vicuna-13B FT         & 8.78  & 4.12  & 1.31 & 0.63 & \textbf{2.72}  \\ \midrule
\textbf{\small Multimodal LLMs} \\
LLaVA-v1.5-7B~\cite{liu2023llava}        & 13.23 & 4.73  & 1.24 & 1.44 & 5.82  \\
LLaVA-v1.5-13B~\cite{liu2023improvedllava}       & {20.56} & 6.07  & 3.44 & 2.85 & {8.19}\\ 
{Gemini-Pro}~\cite{team2023gemini} & 30.98 & 9.05 & 6.81 & 3.66 & 11.46\\
{GPT-4V}~\cite{liu2023improvedllava}       & \textbf{38.72} & 10.53  & 7.14 & 4.03 & \textbf{17.02}\\ 
 \midrule
Human Performance & 82.23 & 0.12 & 0.36 & 1.61 & 80.14\\
\bottomrule
\end{tabular}%
}

\label{tab:results}
\end{wraptable}

% Please add the following required packages to your document preamble:
% \usepackage{booktabs}
% \usepackage{graphicx}
% \begin{table}[]
% \centering
% \caption{Results of GPT-4 and GPT-4V on a subset of 500 samples.}
% \resizebox{0.5\columnwidth}{!}{%
% \begin{tabular}{@{}lcc@{}}
% \toprule
% \textbf{Model}  & \textbf{Sequence Score ($\uparrow$)} & \textbf{Action Score ($\uparrow$)} \\ \midrule
% GPT-4 \cite{openai2023gpt4} &        36.42        &    12.77          \\
% GPT-4V \cite{yang2023dawn} &       \textbf{39.43}         &   \textbf{ 20.76}          \\ \bottomrule
% \end{tabular}%
% }
% \label{tab:ablation}
% \end{table}

As shown in Table~\ref{tab:results}, we experiment with three different categories of models, namely Prompt-based LLMs, Fine-tuned LLMs, and Prompt-based Multimodal Models. 
% GPT-4 is the best-performing model out among all the approaches, which can be attributed to the size of the model, which allows it to score higher on the sequence score and face a lesser penalty for predicting the coordinates and or writing the text. 
GPT-4 is the best-performing approach, scoring higher on the sequence score and invoking lower penalties on coordinate predicting and text input. 
For prompt-only LLMs, the GPT-3.5-turbo and GPT-4 models outperform the other LLM baselines, including the LLaMA~\cite{touvron2023llama} and Vicuna~\cite{vicuna2023} models. We observe that CodeLLaMA-34B~\cite{rozière2023codellama}, which is trained for code generation, also achieves a higher performance than other models of the same size at predicting the action sequences. 

Fine-tuned models also perform much better than their few-shot prompt-only counterparts. Fine-tuning substantially improves LLaMA-13B's sequence score (4.80 to 8.92) and action score (1.62 to 2.14), as well as the other metrics. 
% The training data we use to fine-tune helps the LLMs comprehend the task and learn action sequences more effectively, which is why a major boost in the sequence score is observed. 
Despite this, we observed that both, prompt-based LLMs and finetuned LLMs face severe mouse penalties, especially on click coordinates. This is because they rely solely on text-based signals. 

To address this, we experiment with multimodal language models (Table~\ref{tab:results}). We observe that the coordinate prediction improves significantly when we provide the entire image as input to the multimodal LLM, as this enables it to fully utilize the screen representation. In addition to open sourced models, we also experiment with the GPT-4-vision API~\cite{yang2023dawn} which shows that GPT-4 Vision~\cite{yang2023dawn} outperforms GPT-4 significantly on the Action Score along with improving the sequence score, which we attribute to the strong reasoning abilities of GPT-4 coupled with the improved visual understanding capabilities of the GPT-4-vision model~\cite{yang2023dawn}. These findings pave the way towards exciting new research directions on building multimodal models for long-horizon planning and code generation. \\

\textbf{Human performance over the task:} \ModelName{} consists of visually complicated tasks, and tests various types of computer skills. In order to get a gauge of how well humans perform, we collect evaluation data from human evaluators. We split the test set uniformly amongst 10 human evaluators, and provided them with the screenshot and task instruction. We record the actions taken by the annotators, and measure their performance on our predefined metrics (Table~\ref{tab:results}). 
% We find that users generally perform well on most tasks in their first attempt, but are sometimes unsuccessful in certain tasks owing to factors such as an inability to understand the task, ground the task to the screenshot, or being unfamiliar with the interface of the applications. 
We find that users generally exhibit a high level of proficiency when attempting most tasks for the first time. However, there are instances where users face difficulties in successfully completing certain tasks. These are due to factors including the user's inability to fully comprehend the task, difficulties in grounding the task to the provided screenshot, or a lack of familiarity with the UI.

\section{Conclusion and Future Work}

Autonomous virtual agents offer the potential to automate routine tasks, benefiting users with limited technical expertise. To solve this task, we introduce \ModelName{}, a unique dataset of 9.8K human-labeled data points. \ModelName{} benchmarks autonomous agents across a range of tasks on web and desktop applications. 
% , our groundbreaking and standalone dataset of nearly 10k human-labeled data points, evaluates agents' capabilities in generating executable programs for a range of tasks across websites and desktop applications. 
LLM-based agents, like GPT-4, achieve a respectable action score of 11.6 on our dataset. However, \ModelName{} presents a challenge for the current state-of-the-art language and multimodal models. It provides a direction for future research on foundational multimodal models that seamlessly integrate language and visual understanding of computer screens and stands poised to drive the next wave of advancements in generalist autonomous agents offering omnipotent assistance to humans$\text{.}$

\section{Limitations}

This work introduces a valuable dataset, yet we recognize a few limitations that exist. State-of-the-art models like GPT-4, may exhibit susceptibility to hallucinations and bias towards specific data types, hindering broad applicability. Reliance on closed models like GPT-4V poses integration challenges due to high costs and time constraints. Despite efforts for equal representation and data collection without personal information, biases may be introduced as the dataset is exclusively in English, and human-curated content may have temporal biases.

\section{Ethics Statement}
As a part of the dataset creation process, we carefully review the pipeline at every stage, ensuring there is no personally identifiable information or offensive content, either during data collection or through the use of LLMs. For all purposes, we create dummy accounts that mimic real-like user content. To get the gold labels scripts we seek help from well-qualified student workers, approved through the institution, and get the bounding box data annotated through MTurk workers, both of whom are paid $\$25$ per hour, which is greater than the minimum wage rate (We detail this process in the supplementary material). Human studies are also done with the help of student workers approved by the institution at the above-mentioned payscale. We also ensure that all groups have equitable representation and that no personal opinions are reflected in the dataset, avoiding bias during the collection as well as the annotation process.

\section*{Acknowledgements} We extend our heartfelt gratitude to Writer.com for their generous and unwavering support throughout this project. Their dedicated team's expertise and collaboration were invaluable in achieving our goals. 

% ---- Bibliography ----
%
% BibTeX users should specify bibliography style 'splncs04'.
% References will then be sorted and formatted in the correct style.
%
\bibliographystyle{splncs04}
\bibliography{main}

\end{document}